\documentclass[11pt,a4paper]{article}
\usepackage[a4paper,margin=1in]{geometry}
\usepackage[utf8]{inputenc}
\usepackage[protrusion=false,expansion=false]{microtype}
\usepackage{graphicx}
\usepackage{booktabs}
\usepackage{array}
\usepackage{amsmath,amssymb}
\usepackage{xcolor}
\usepackage[colorlinks=true,linkcolor=blue!50!black,citecolor=teal!50!black,urlcolor=blue!60!black]{hyperref}
\usepackage{natbib}
\usepackage{enumitem}
\usepackage{caption}
\usepackage{tikz}
\usetikzlibrary{positioning,arrows.meta,calc,fit,shapes.geometric,backgrounds}
\graphicspath{{./}{figures/}{../figures/}{../figures_codex_independent/}{../../figures_codex_independent/}}
\newcolumntype{L}[1]{>{\raggedright\arraybackslash}p{#1}}

\title{\bf More Than Memory:\\
\large Task-Conditioned Signed FFN Writes in Long-Context Retrieval}

\author{Zhibo Yang\\
Ocean University of China\\
\texttt{yzbdeeplearning@163.com}}
\date{June 2026}

\begin{document}
\maketitle

% =============================================================================
\begin{abstract}
% =============================================================================
FFNs are often treated as parametric memories. In long-context retrieval, however, the
sharper question is not only what they store, but whether their native residual writes push
the current retrieval state toward or away from the correct answer. We test this by scaling
the model's own FFN write one layer at a time, without editing weights or injecting
external steering vectors.

Across controlled literal and semantic retrieval suites, native FFN response surfaces are
signed, layer-specific, and task-conditioned: the final FFN is a suppressor in 7 of 8
model--suite cases, and 60\% of layers switch role between retrieval modes (95\% CI
[50\%, 69\%]). A local directional derivative along the native write separates the two
monotone roles: suppressors have negative derivative in 34/35 cases, and amplifiers have
positive derivative in 18/18 cases, so the roles are not reducible to write size. On a
safety-filtered LongBench retrieval-QA probe, the same diagnostic predicts attenuation
damage with raw $R^2{=}0.796$ on Qwen2.5-7B and $0.791$ on Qwen3.5-9B; a held-out
suppressor-attenuation policy improves retrieval margins over random and norm-matched
controls. These results show that native FFN scaling exposes a signed, task-conditioned
residual-write structure in retrieval, and that write-gradient alignment is a compact
diagnostic for the two monotone roles.
\end{abstract}

% =============================================================================
\section{Introduction}
% =============================================================================
\label{sec:intro}

A central result of mechanistic interpretability is that Transformer feed-forward networks
(FFNs) behave like key-value memories: they store associations, enrich residual states with
factual information, and expose editable factual content
\citep{geva2021transformer,geva2022transformer,meng2022rome,dai2022knowledge,geva2023dissecting}.
This view now has engineering consequences. Recent architectures make memory more explicit,
sparse, or modular: Memory Layers add trainable key-value lookup modules
\citep{berges2024memorylayers}; DeepSeek Engram introduces conditional memory lookup as a
separate sparsity axis \citep{cheng2026engram}; MemoryLLM decouples FFN memory from
self-attention and residual flow \citep{jaiswal2026memoryllm}; STEM replaces part of FFN
computation with embedding modules \citep{sadhukhan2026stem}; and PEER-style routing
retrieves from many tiny experts \citep{he2024millionexperts}. The operational premise is
that FFN memory can be factored out, sparsified, or replaced without losing the relevant
capability.

\begin{figure}[h]
\centering
\includegraphics[width=0.50\linewidth]{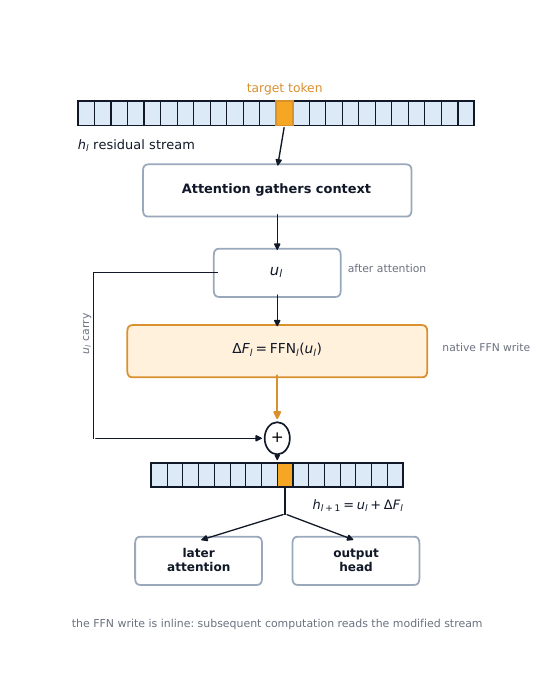}
\caption{\textbf{Native FFN residual flow without intervention.} Attention first
aggregates contextual evidence into the target-token residual state, the FFN computes a
native write $\Delta F_l$, and residual addition writes
$h_{l+1}=u_l+\Delta F_l$ into the stream consumed by later attention layers and the output
head. No gain scaling or damage metric is applied here.}
\label{fig:native-flow}
\end{figure}

The architectural complication, illustrated in Figure~\ref{fig:native-flow}, is that a
native FFN is not a side table. The point is not that FFNs are the only Transformer
modules that operate on the residual stream; attention and normalization also participate
in residual computation. The point is narrower and more consequential for the memory view:
the memory-like FFN computation is written inline into the same state that later retrieval
computation consumes.

A Transformer layer writes
\[
 h_{l+1}=u_l+\Delta F_l,\qquad
 u_l := h_l+\mathrm{Attn}_l(h_l),
\]
into a residual stream that every subsequent layer reads. In long-context retrieval,
attention first aggregates contextual evidence into the target-token residual state; the
FFN then applies the only token-wise nonlinear transformation between consecutive attention
layers and writes $\Delta F_l$ back into the same stream; later attention layers and the
output head consume that transformed state. This raises two questions: \textbf{Q1}, does an
FFN write affect cross-layer retrieval information flow? \textbf{Q2}, if it does, in which
direction, and is that direction task-dependent?

We test this view with native FFN scaling (Figure~\ref{fig:concept_write}):
\[
 h_{l+1}\leftarrow u_l+\alpha\,\Delta F_l,\qquad \alpha\in\{0,0.25,\ldots,2\},
\]
one layer at a time. We use this intervention because it isolates the causal role of the
native residual write. It does not edit weights, inject an external steering vector, or
replace the attention pattern; it preserves the learned content and direction of
$\Delta F_l$ while varying only its gain at a single layer. The identity point
$\alpha{=}1$ is the unmodified model, $\alpha{<}1$ asks whether attenuating or removing the
same write helps retrieval, and $\alpha{>}1$ asks whether stronger expression of the same
write amplifies its effect. The resulting dose-response therefore distinguishes two
accounts. If the FFN acted only as passive memory, reducing its write should mainly remove
useful stored information. The signed-write hypothesis predicts a stronger and more specific
pattern: the response should be signed. Some layers should help retrieval when skipped,
some should hurt when skipped, and the same layer can change role when the retrieval mode
changes.

The forward intervention answers whether FFN writes have signed causal effects. The next
question is whether the pattern can be explained from architecture-internal signals. Around
the native point $\alpha{=}1$, the retrieval margin changes along the model's own FFN
write direction. At layer $l$, the forward FFN write $\Delta F_l$ and the backward
retrieval gradient $\nabla M$ meet at the same residual node
(Figure~\ref{fig:concept_diagnostic}). Their inner product
\[
 s_l=\Delta F_l^{\!\top}\nabla M
\]
is the local directional derivative of retrieval margin along the native write. We use this
quantity as a mechanism anchor, not as a substitute for the causal intervention: the
scaling experiment defines the role, and the directional derivative tests whether the role
follows the write direction.

\paragraph{Contributions.}
\begin{enumerate}[leftmargin=*,itemsep=2pt]
\item \textbf{Signed FFN role taxonomy.} Native FFN scaling reveals calibrated, amplifying, and
 suppressive roles in long-context retrieval. These roles are layer-specific and
 task-conditioned: the final layer is a suppressor in 7/8 controlled model--suite cases,
 and 60\% of layers switch role between literal and semantic retrieval.
 \item \textbf{Directional mechanism and controls.} The local directional derivative anchors the two
 monotone roles: suppressor writes point against the retrieval objective in 34/35 cases,
 while amplifier writes point with it in 18/18 cases. Seed stability, threshold sweeps,
 and residual-size controls show that the taxonomy is not a seed artifact, threshold
 artifact, or norm-only effect.
 \item \textbf{Naturalistic transfer and diagnostic implication.} A safety-filtered LongBench
 retrieval-QA probe transfers the same diagnostic to natural QA, and a targeted attenuation
 sanity check shows that suppressor layers are causally useful control points. This gives a
 concrete diagnostic for FFN replacement or decoupling studies in retrieval settings:
 evaluate the write direction, not only lookup accuracy.
\end{enumerate}

\begin{table}[h]\centering\scriptsize
\caption{\textbf{Claim--test contract.} The paper is organized as a single chain:
signed causal effect, task conditioning, and local directional explanation.}
\label{tab:claimtest}
\begin{tabular}{L{0.20\linewidth}L{0.24\linewidth}L{0.29\linewidth}L{0.09\linewidth}}
\toprule
Claim & Test & What would falsify it & Section \\\midrule
Native FFN writes have signed causal effects, not only unsigned importance. &
Layerwise native scaling; full-alpha heatmaps; endpoint taxonomy. &
All non-identity gains mainly degrade retrieval, or layers differ only in magnitude rather than sign. &
\S\ref{sec:c1} \\
Signed roles are task-conditioned. &
Literal/semantic contrast; same-layer role switches and opposite-sign examples. &
Most layers keep the same role across retrieval modes, except near-zero boundary cases. &
\S\ref{sec:c2} \\
Monotone roles follow local write direction. &
Role $\times$ sign cross-tab; first-order attenuation prediction; norm/depth controls. &
Suppressor/amplifier signs are mixed, or simple norm/depth features explain roles as well. &
\S\ref{sec:c3}, \S\ref{sec:robust} \\
The diagnostic transfers beyond synthetic retrieval. &
Safety-filtered LongBench retrieval-QA probe on Qwen models. &
The signed surface disappears, or the local directional predictor loses sign/rank signal. &
\S\ref{sec:longbench} \\\bottomrule
\end{tabular}
\end{table}

% =============================================================================
\section{Related Work}
% =============================================================================
\label{sec:related}

\subsection{FFN as memory and its engineering consequences}
\citet{geva2021transformer,geva2022transformer} establish the FFN-as-memory view by
interpreting FFN sublayers as key-value memories. Editing work shows that factual
associations are localizable at FFN weights \citep{meng2022rome,dai2022knowledge}, and
\citet{geva2023dissecting} trace factual enrichment to early MLP sublayers with final
extraction mediated by attention. This line explains what FFNs store and how stored
associations can be localized or edited.

Recent architecture work turns this memory view into a design axis. Memory Layers add
explicit trainable lookup modules \citep{berges2024memorylayers}; DeepSeek Engram adds
conditional memory lookup as a separate sparsity axis \citep{cheng2026engram}; MemoryLLM
decouples FFN memory from self-attention and residual flow \citep{jaiswal2026memoryllm};
STEM replaces part of FFN computation with embedding modules \citep{sadhukhan2026stem};
and PEER-style routing retrieves from many tiny experts \citep{he2024millionexperts}.
These works decouple memory architecturally. We diagnose a complementary property in
conventional Transformers: the memory-like FFN computation is written inline into the
residual stream. The decoupling literature addresses \emph{what} the module retrieves; we
ask \emph{where and in which direction} the retrieved or transformed vector pushes the
current retrieval objective. \S\ref{sec:discussion} discusses the narrower design
consequence for retrieval settings: FFN replacement studies should check whether the
replacement preserves retrieval-relevant write direction, not only lookup behavior.

\subsection{Layer-wise FFN importance}
\citet{pires2023onewide} show that FFNs can be redundant and replaceable by one widened FFN;
\citet{men2024shortgpt} rank blocks by redundancy; and \citet{ikeda2025layerwise} show that
reallocating FFN capacity to middle layers can improve downstream tasks. These works report
FFN importance as a magnitude, redundancy, or relative-benefit score. Our role taxonomy is
complementary: it measures a \emph{signed}, \emph{task-dependent} per-layer response. A
large FFN write is not necessarily helpful, and a small write is not necessarily irrelevant;
the direction of the native write relative to the retrieval objective determines the two
monotone roles.

\subsection{Long-context retrieval}
Long-context benchmarks show that retrieval is behaviorally subtle
\citep{kamradt2023niah,liu2023lost,hsieh2024ruler,bai2024longbench,zhang2024infinitebench}.
Most relevant is \citet{modarressi2025nolima}, which shows that literal and semantic
retrieval are distinct capabilities and attributes the gap primarily to attention. Our
experiments localize part of that divergence to the FFN write direction: the same FFN layer
can switch its entire causal role when the retrieval mode changes. LongBench is used here
not as a leaderboard benchmark but as a natural-language retrieval-QA probe for testing
whether the signed surface and directional-derivative diagnostics transfer beyond controlled
synthetic retrieval.

\subsection{Activation steering}
\citet{subramani2022steering,turner2023actadd,li2023iti,panickssery2024caa} steer behavior
by injecting residual vectors; \citet{bas2025steering} report an inverted-U in steering
coefficient. We instead scale the model's \emph{native FFN residual write} and read the
response as a signed role taxonomy. The contribution is not that residual interventions can
change behavior; it is that native FFN writes form a task-conditioned signed structure, and
that the monotone roles are anchored by the local directional derivative of retrieval
margin along the native write.

% =============================================================================
\section{Native FFN Writes and Local Direction}
% =============================================================================
\label{sec:hypothesis}

Figure~\ref{fig:native-flow} shows where the native FFN write enters the residual stream before
intervention. Figure~\ref{fig:concept_write} summarizes the forward residual-write intervention.
Figure~\ref{fig:concept_diagnostic} summarizes the endpoint-response and
directional-derivative diagnostics used to classify and explain the signed roles.

{\captionsetup{hypcap=false}
\begin{center}
\centering
\includegraphics[width=0.52\linewidth]{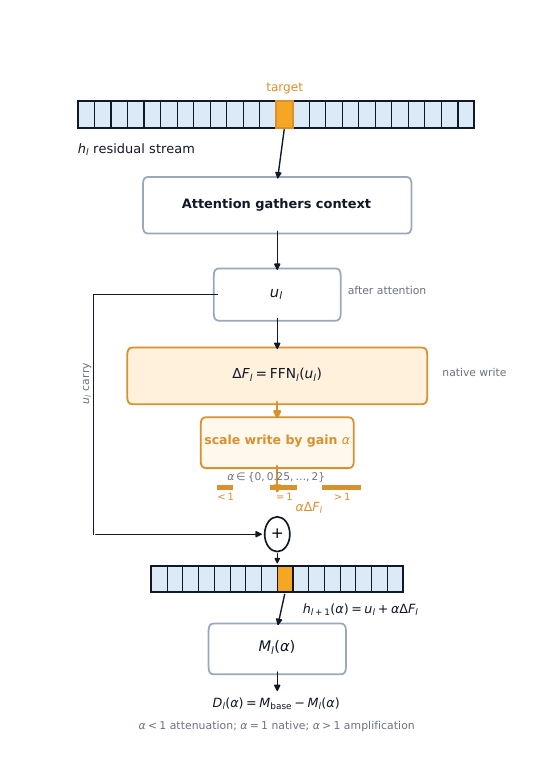}
\captionof{figure}{\textbf{FFN residual-write intervention.} In a normal forward pass,
attention aggregates contextual evidence into the target-token residual state, and the FFN
writes a transformed vector $\Delta F_l$ back into the same residual stream that later
attention reads. Native FFN scaling changes only the strength of that write,
$h_{l+1}\leftarrow u_l+\alpha\Delta F_l$, and the readout margin gives
$D_l(\alpha)=M_{\mathrm{base}}-M_l(\alpha)$.}
\label{fig:concept_write}
\end{center}
}

{\captionsetup{hypcap=false}
\begin{center}
\centering
\includegraphics[width=0.68\linewidth]{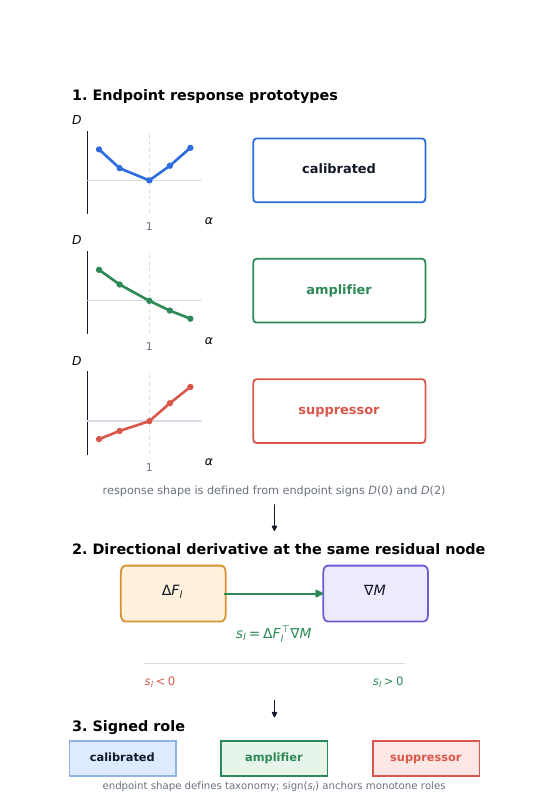}
\captionof{figure}{\textbf{Signed-role diagnostic.} Endpoint responses classify each
layer by the signs of $D_l(0)$ and $D_l(2)$, yielding calibrated, amplifying, or
suppressive roles. The backward diagnostic measures whether the same native write points with or
against the retrieval objective: the forward write $\Delta F_l$ and retrieval gradient
meet at the same residual node, and the inner product
$s_l=\Delta F_l^{\!\top}\nabla M$ is the local directional derivative anchoring the two
monotone roles.}
\label{fig:concept_diagnostic}
\end{center}
}

\subsection{Architectural fact: native FFN writes are consumed downstream}
A pre-norm Transformer layer evolves the residual stream as
\[
 h_{l+1}=h_l+\mathrm{Attn}_l+\mathrm{FFN}_l=u_l+\Delta F_l .
\]
In Llama-, Qwen-, and SmolLM2-style sequential models, $\mathrm{FFN}_l$ reads $u_l$; in
Pythia- and GPT-NeoX-style parallel models, it reads $h_l$ (\S\ref{sec:limitations}). In retrieval,
attention aggregates contextual evidence into the target-token residual state. The FFN then
applies a token-wise nonlinear transformation and writes $\Delta F_l$ back into the same
residual stream. Later attention layers and the output head consume this transformed state
(Figure~\ref{fig:concept_write}). The residual connection itself is not a new observation. What
we test is whether the \emph{direction} of $\Delta F_l$ is signed, task-conditioned, and
predictable from the local directional derivative.

\subsection{Intervention design: native FFN scaling}
We scale the native write
\[
 h_{l+1}\leftarrow u_l+\alpha\Delta F_l,\qquad
 \alpha\in\{0,0.25,\ldots,2\},
\]
one layer at a time (Figure~\ref{fig:concept_write}). We report
$D_l(\alpha)=M_{\mathrm{base}}-M_l(\alpha)$, so $D_l(\alpha)>0$ means the intervention
hurt retrieval and $D_l(\alpha)<0$ means it helped. We use endpoints rather than AUC or
curve slope because the role taxonomy is meant to summarize two explicit causal probes:
removing the native write ($\alpha=0$) and over-amplifying it ($\alpha=2$). The
$|D_l(\alpha)|<0.1$ weak band is a pre-specified margin tolerance for near-zero endpoint
responses; \S\ref{sec:robust} and Appendix~\ref{app:thresholds} report the corresponding
threshold sweep. A layer's endpoint role is defined by the signs at $\alpha=0$ and
$\alpha=2$:
\begin{itemize}[leftmargin=*,itemsep=1pt]
 \item \textbf{calibrated}: skip hurts and amplify hurts; the pretrained gain is locally
 calibrated.
 \item \textbf{amplifier}: skip hurts and amplify helps; more FFN write improves retrieval.
 \item \textbf{suppressor}: skip helps and amplify hurts; more FFN write damages retrieval.
 \item \textbf{weak/mixed}: either endpoint lies inside the $|D_l(\alpha)|<0.1$ weak band or
 the endpoint pattern is not one of the three roles above.
\end{itemize}
Native scaling is not intended to preserve residual norm; it is a causal dose intervention
on the model's own FFN write. Norm-preserving directional controls are used separately to
test whether the signed effect is reducible to update magnitude
(\S\ref{sec:robust}, Appendix~\ref{app:fixednorm}).
This intervention tests two predictions. \textbf{P1}: native FFN writes produce signed,
layer-structured response surfaces rather than uniform degradation. \textbf{P2}: the same
layer's role can switch between literal and semantic retrieval because the task changes what
counts as useful evidence at the residual state.

\subsection{Local directional derivative and backward mechanism}
The forward intervention identifies the response surface; the directional criterion explains
the two monotone roles. Let
\[
 M_l(\alpha)=M(u_l+\alpha\Delta F_l)
\]
denote the retrieval margin after scaling the native FFN write at layer $l$. Around the
unmodified model, $\alpha{=}1$, a first-order expansion gives
\[
 M_l(\alpha)-M_l(1)\approx(\alpha-1)\Delta F_l^{\!\top}\nabla M .
\]
Thus
\[
 D_l(\alpha)=M_l(1)-M_l(\alpha)\approx (1-\alpha)s_l,\qquad
 s_l:=\Delta F_l^{\!\top}\nabla M .
\]
We interpret $s_l$ as the local directional derivative of retrieval margin along the
native FFN write. The first-order formula itself is standard; the empirical claim is that
this simple local quantity remains informative across layers, models, retrieval modes,
non-infinitesimal attenuation gains, and LongBench QA transfer, while magnitude-only
controls do not explain the same role structure. This is a diagnostic, not a second causal
intervention: the causal object is still the native write $\Delta F_l$ under scaling. If
$s_l>0$, the write locally points toward higher retrieval quality and should behave as an
amplifier under attenuation; if $s_l<0$, the write points against retrieval and should
behave as a suppressor. Calibrated layers are dose-shape roles and are not claimed to be
sign-determined.

This local interpretation has a limited scope. It most directly constrains attenuation-side
signs and ranks near the native point; it need not exactly fit damage amplitude across all
gains, task protocols, or depths. \textbf{P3}: the two monotone roles are predicted by the
sign of $s_l$ (suppressor$\Rightarrow s_l<0$, amplifier$\Rightarrow s_l>0$), and the same
quantity gives the zero-parameter local prediction
$D_l^{\mathrm{pred}}(\alpha)=(1-\alpha)s_l$ for attenuation damage.

% =============================================================================
\section{Experimental Setup}
% =============================================================================
\label{sec:setup}
\paragraph{Metric: retrieval damage $D_l(\alpha)$.} Throughout, our dependent variable is the
\emph{retrieval margin} of the gold answer: the mean logit margin (gold logit minus the
best non-gold logit) over the retrieval task's answer tokens. The intervention effect is
reported as $D_l(\alpha):=M_{\mathrm{base}}-M_l(\alpha)$
(\texttt{raw\_damage\_margin}), so $D_l(\alpha)>0$ means the intervention \emph{hurt}
retrieval (lowered the gold margin), and $D_l(\alpha)<0$ means it \emph{helped}. Identity
($\alpha{=}1$) gives $D_l(1){=}0$ by construction. Per-condition $D_l(\alpha)$ is averaged over (i) three random seeds for the
three main-protocol models and the available SmolLM2 cross-protocol validation run, and (ii) three
difficulty levels within each suite: $L_1$ (single key-value), $L_2$ (needle at varying
positions), $L_3$ (conflicting distractors). Bootstrap confidence intervals (over seeds for
the main-protocol end-layer endpoints; over layers for the role-switch rate) are reported
in \S\ref{sec:robust}.

\paragraph{Models.} Three main-protocol models span $\sim$1--9B and both
residual-composition families (Table~\ref{tab:models}): Pythia-1B
\citep{biderman2023pythia} (parallel), Qwen2.5-7B \citep{qwen25} (sequential), and
Qwen3.5-9B-Base (sequential). Qwen3.5-9B-Base denotes a Qwen-family sequential 32-layer
checkpoint used as an additional mechanism-validation checkpoint in our runs; we report it
separately from the standard public Qwen2.5 release and do not use it as a public benchmark
reference. Results are reported per model rather than merged into a leaderboard-style score.
SmolLM2-1.7B \citep{allal2025smollm2} provides external
architecture and data-protocol validation. All
four have complete literal and semantic suites. Backward-gradient diagnostics (bf16
full-alpha) are available for the three main-protocol models and for sampled SmolLM2
anchor layers; all four are subjects of \S\ref{sec:c3}.

\paragraph{Cross-protocol validation design.} The model set separates estimation from
stress testing. Pythia-1B, Qwen2.5-7B, and Qwen3.5-9B form the unified, three-seed
main-protocol set used for seed aggregation, leave-one-seed-out stability, and the primary
Qwen response-surface figures. SmolLM2 changes both model family and data protocol relative
to those three models, providing a cross-protocol stress test. Its literal suite matches
the final-layer suppressor pattern; its semantic suite yields a calibrated role, showing how the
same taxonomy records protocol-sensitive role changes. We report SmolLM2 side by side for the taxonomy,
task-conditioning, block-depth pattern, and sampled backward-anchor role/sign relation,
while keeping seed-LOO summaries restricted to the three main-protocol models.

\begin{table}[h]\centering\small
\caption{Models and residual composition.}\label{tab:models}
\begin{tabular}{lrlr}\toprule
Model & Params & Residual composition & \# layers \\\midrule
Pythia-1B & 1.0B & parallel (GPT-NeoX) & 16 \\
SmolLM2-1.7B & 1.7B & sequential (Llama) & 24 \\
Qwen2.5-7B & 7B & sequential & 28 \\
Qwen3.5-9B-Base & 9B & sequential & 32 \\\bottomrule
\end{tabular}\end{table}

\paragraph{Retrieval suites.} \textbf{Literal}: verbatim needle, question shares vocabulary
(NIAH lineage \citep{kamradt2023niah}). \textbf{Semantic}: needle and question have minimal
lexical overlap, forcing associative retrieval (NoLiMa-style \citep{modarressi2025nolima}).
Both at 8K context, three difficulty levels ($L_1$/$L_2$/$L_3$).

\paragraph{Naturalistic LongBench retrieval-QA probe.} To evaluate the signed surface under
natural-language long-context QA, we add a safety-filtered LongBench retrieval-QA probe
\citep{bai2024longbench} on Qwen2.5-7B and Qwen3.5-9B. We use two candidate-scored
LongBench subsets, Qasper and MultiFieldQA-en, with 100 examples each (200 total), seed 13,
bf16, and the same native FFN scaling grid $\alpha\in\{0,0.25,\ldots,2\}$. Qwen2.5 has a
complete $28{\times}9$ layer--gain grid plus baseline (253 summary rows, 50{,}600
predictions); Qwen3.5 has a complete $32{\times}9$ grid plus baseline (289 summary rows,
57{,}800 predictions) with tail-shard merge validation. We keep this probe separate
from the three-seed controlled taxonomy: it evaluates transfer of the signed surface and
directional-derivative mechanism under a natural QA protocol.

\paragraph{LongBench subset selection.} LongBench is used here as natural-language
long-context retrieval validation, not as a leaderboard benchmark. We therefore select
subsets by mechanism-experiment constraints. A subset must be QA/retrieval-shaped, have
answers that can be localized or inferred from the context, support forced-choice logprob
scoring with same-subset distractors, and use relatively short answer spans so that the
measured margin is not dominated by generation format, output length, or fluency. We also
prefer English natural-language contexts, avoid code, summarization, dialogue,
classification, and Chinese subsets in this first probe, and avoid strong multi-hop
reasoning so that the task primarily probes retrieval/readout rather than complex
reasoning. We further apply a conservative subtask-level safety screen and include only QA
subtasks with zero strict NSFW hits in the strict audit. Qasper and MultiFieldQA-en satisfy
these constraints while providing real long-form natural-language contexts and enough
answer-pool diversity for length-matched distractors.

\paragraph{Backward diagnostic filtering.} Qwen supplementary backward runs include a
short-copy control alongside the semantic tasks. For semantic directional-derivative
diagnostics, we average only the semantic tasks ($S_1$--$S_3$) and exclude that control.
This affects the directional-derivative diagnostic but not the endpoint taxonomy, which is
computed from the intervention suites.

\paragraph{Provenance.} All headline numbers are re-verified from raw per-run summaries or
per-sample prediction files. The accompanying project artifacts record the reconstruction
scripts, audit notes, LongBench merge checks, and robustness tables used to produce the
paper figures and tables.

\paragraph{Sign convention.} Throughout, positive $D_l(\alpha)$ indicates retrieval damage
and negative $D_l(\alpha)$ indicates retrieval improvement.

% =============================================================================
\section{Results}
% =============================================================================
\label{sec:results}

\subsection{Q1: Native FFN scaling reveals signed causal response surfaces}
\label{sec:c1}

\begin{figure}[h]
\centering
\includegraphics[width=\linewidth]{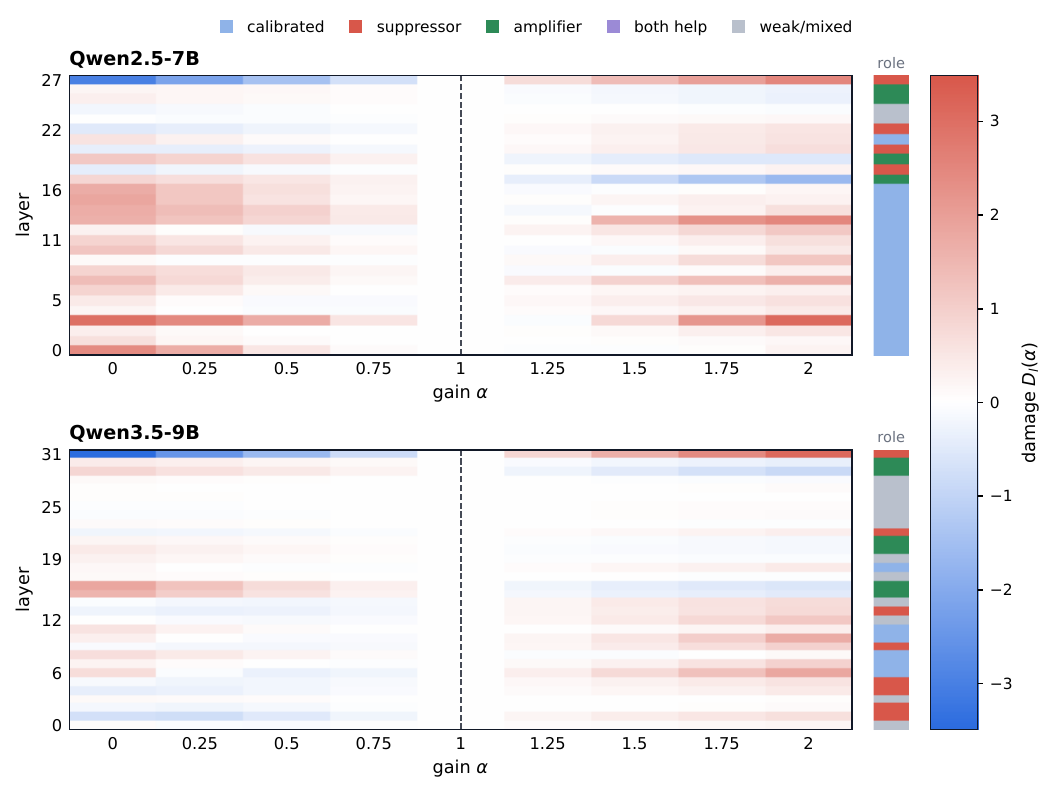}
\caption{\textbf{Qwen FFN response surfaces.}
Seed-averaged literal-retrieval damage $D_l(\alpha)$ across layers and gains. The dashed
column is $\alpha{=}1$; side strips show endpoint roles.}
\label{fig:qwen-surface}
\end{figure}

Figure~\ref{fig:qwen-surface} shows the full-alpha response surface for the two Qwen
models: every layer is probed across gains, and the color records signed retrieval damage
($D_l(\alpha){>}0$ hurts, $D_l(\alpha){<}0$ helps). The important fact is visible before any
summary statistic: FFN response is not a scalar importance curve. Adjacent layers can have
opposite signs under the same gain, and Qwen2.5-7B and Qwen3.5-9B have visibly different
surfaces. Qwen2.5 literal retrieval contains many calibrated layers; Qwen3.5 literal retrieval
has a mixed surface with suppressor, amplifier, and weak/mixed regions.
A generic perturbation-sensitivity account would predict degradation under gain changes;
the heatmaps instead show structured, signed variation. The final-layer suppressor cases
are also matched by negative $s_l$ in the directional-derivative diagnostic, a property
that generic fragility does not predict.

The endpoint taxonomy in Table~\ref{tab:rolecounts} extends the same conclusion across
models and retrieval modes. In the three main-protocol models, all six final FFN layers are
suppressors: skipping them helps retrieval, while amplifying them hurts. SmolLM2 then acts
as the intended cross-protocol validation: its literal suite matches the final-suppressor
pattern, while its semantic suite is calibrated. This is the point of including an
external protocol. The result separates the robust taxonomy claim from the final-layer
specialization: signed roles persist under protocol shift, and the semantic final layer
moves into a calibrated role.

\begin{table}[h]\centering\small
\caption{\textbf{Endpoint taxonomy counts.} Roles are computed from seed-averaged endpoint
damage $D_l(0)$ and $D_l(2)$ with threshold $|D_l(\alpha)|{>}0.1$. Rare both-help and
below-threshold endpoint patterns are folded into weak/mixed.}
\label{tab:rolecounts}
\begin{tabular}{llccccc}\toprule
Model & Suite & Calib. & Supp.\ & Amp.\ & Weak/mixed & Final role \\\midrule
Pythia-1B & literal & 2 & 5 & 2 & 7 & suppressor \\
 & semantic & 8 & 3 & 0 & 5 & suppressor \\
SmolLM2-1.7B & literal & 10 & 7 & 3 & 4 & suppressor \\
 & semantic & 12 & 1 & 4 & 7 & calibrated \\
Qwen2.5-7B & literal & 18 & 4 & 4 & 2 & suppressor \\
 & semantic & 14 & 4 & 3 & 7 & suppressor \\
Qwen3.5-9B & literal & 6 & 8 & 6 & 12 & suppressor \\
 & semantic & 9 & 9 & 2 & 12 & suppressor \\\bottomrule
\end{tabular}
\end{table}

\begin{figure}[h]
\centering
\includegraphics[width=0.90\linewidth]{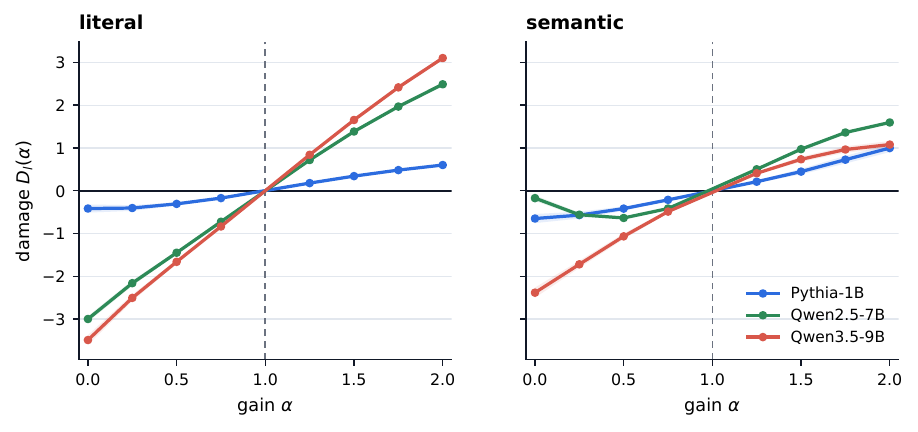}
\caption{\textbf{Final FFN layers are usually readout-adjacent suppressors.}
In the three main-protocol models, both literal and semantic final FFNs show the same
qualitative pattern: $\alpha{<}1$ helps and $\alpha{>}1$ hurts. Qwen3.5-9B literal is the
largest case ($\alpha{=}0\to{-}3.49$, $\alpha{=}2\to{+}3.10$). SmolLM2 semantic is
calibrated, showing that the same taxonomy records protocol-sensitive final-layer roles.}\label{fig:endlayer}
\end{figure}

\subsection{Q2: Signed FFN roles are task-conditioned}
\label{sec:c2}

Literal and semantic retrieval differ in the controlled dimension that matters for this
paper: lexical overlap. If an FFN layer were simply a static store for the target fact, its
role should not systematically change when the retrieval mode changes. It does. Across
the 100 layers with both-suite endpoint data, 60 switch role between literal and semantic
retrieval (60\%, 95\% CI [50\%, 69\%] by layer bootstrap). At the sharper sign level,
using attenuation at $\alpha{=}0.25$ with $|D_l(\alpha)|{>}0.1$ in both suites, we find 3
opposite-sign layers in Pythia-1B, 4 in Qwen2.5-7B, 6 in Qwen3.5-9B, and 7 in the
SmolLM2 cross-protocol run.

\begin{figure}[h]
\centering
\includegraphics[width=0.86\linewidth]{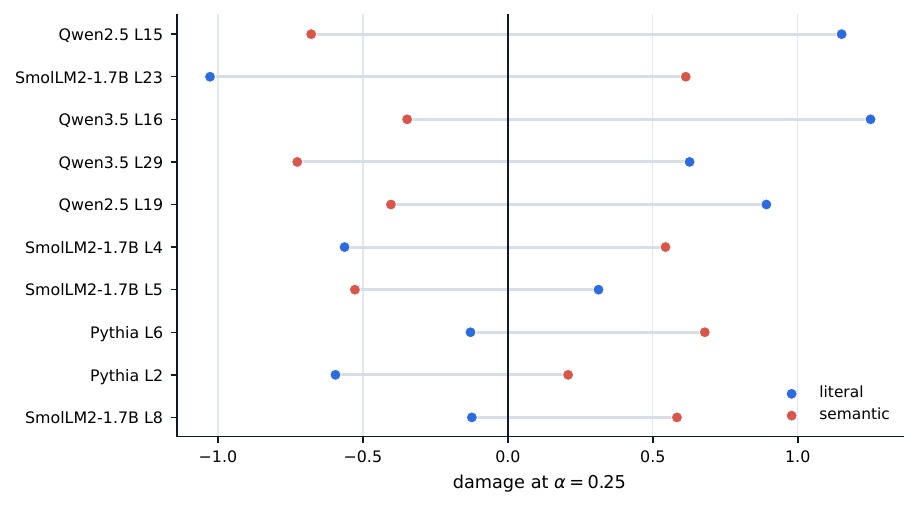}
\caption{\textbf{Same-layer FFN effects can flip between retrieval modes.}
Top opposite-sign examples at $\alpha{=}0.25$. Blue bars are literal retrieval; red bars
are semantic retrieval. Positive damage means the intervention hurt retrieval.}
\label{fig:task}
\end{figure}

These are not small threshold artifacts around zero. In Qwen2.5-7B layer 15, weakening the
FFN hurts literal retrieval but helps semantic retrieval; in Qwen3.5-9B layer 29, the same
pattern appears with comparable magnitude in both directions. Under the signed-write
view, the same layer's task gradient can rotate, changing the sign of
$s_l=\Delta F_l^\top\nabla M$ and reassigning the layer's role. We verify that mechanism
anchor in \S\ref{sec:c3}. Semantic retrieval also exposes a propagation/amplitude
component, so the role switch analysis combines the local direction signal with the
downstream propagation diagnostics summarized in Appendix~\ref{app:block}.

\subsection[Q3: local directional derivative]{Q3: Local directional derivative explains monotone roles}
\label{sec:c3}

The mechanism test is targeted. We compute
$s_l=\Delta F_l^{\!\top}\nabla M$ from backward-gradient diagnostics, then ask whether the
sign of $s_l$ predicts the behavioral endpoint role. The target of this test is the two
\emph{monotone} roles --- suppressor and amplifier --- because they are single-direction
effects. The \emph{calibrated} role is a different dose shape: both endpoints hurt, so it is
modeled as a calibrated-gain role rather than a sign role.

\begin{figure}[h]
\centering
\includegraphics[width=\linewidth]{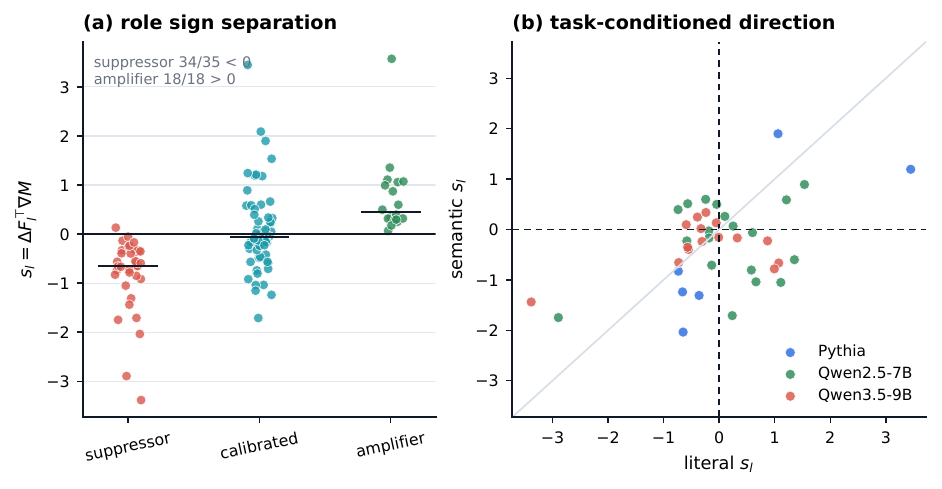}
\caption{\textbf{Suppressor/amplifier roles are predicted by the sign of
$s_l=\Delta F_l^{\!\top}\nabla M$}. Across the three full-backward main-protocol models
plus SmolLM2 backward anchor layers, suppressor layers have
$s_l{<}0$ in 34/35 cases, and amplifier layers have $s_l{>}0$ in 18/18 cases. Calibrated
layers are shown as the non-monotone comparison class. The right panel
illustrates that the same layer's $s_l$ can flip between literal and semantic retrieval,
giving a mechanism instance of the role switches.}\label{fig:sl}
\end{figure}

Figure~\ref{fig:sl} and Table~\ref{tab:sl-role} give the main mechanism anchor. Across the
three full-backward main-protocol models plus SmolLM2 backward anchor layers, suppressor
layers have $s_l{<}0$ in 34/35 cases (pooled mean $-0.81$), and
amplifier layers have $s_l{>}0$ in 18/18 cases (pooled mean $+0.75$). The single suppressor
exception is Qwen3.5 semantic layer 2 after excluding the short-copy control from the
semantic backward diagnostic. The pattern is otherwise model-local: Pythia suppressor 8/8,
amplifier 2/2; Qwen2.5 suppressor 8/8, amplifier 7/7; Qwen3.5 suppressor 16/17, amplifier
8/8; SmolLM2 suppressor 2/2, amplifier 1/1. Calibrated layers are balanced around sign (27/61
positive), matching their interpretation as calibrated-gain layers rather than monotone
direction layers.

\begin{table}[h]\centering\small
\caption{\textbf{Role $\times$ $s_l$ sign cross-tab.} Counts pool the three full-backward
main-protocol models plus SmolLM2 backward anchor layers.
$n{=}161$ layer-suite pairs: Pythia-1B 32 (16 layers $\times$ 2 suites), Qwen2.5-7B 56,
Qwen3.5-9B 64, SmolLM2-1.7B 9 (its 9 backward anchor layers, literal suite only).
Near-perfect separation for the two monotone roles; calibrated is sign-uninformative.}
\label{tab:sl-role}
\begin{tabular}{lrrrr}\toprule
Role & $n$ & $s_l{<}0$ & $s_l{>}0$ & mean $s_l$ \\\midrule
suppressor  & 35 & 34 & 1 & $-0.81$ \\
amplifier   & 18 & 0  & 18 & $+0.75$ \\
calibrated & 61 & 34 & 27 & $+0.07$ \\
other       & 47 & 26 & 21 & $-0.04$ \\\bottomrule
\end{tabular}\end{table}

\paragraph{First-order prediction.} If $s_l$ is the signed contribution per unit
$\alpha{-}1$, then $D_l(\alpha)\approx -s_l\,(\alpha{-}1)$ at small $|\alpha{-}1|$. We
test this zero-parameter predictor on each model separately at attenuation alphas
$\{0.25,0.5,0.75\}$. In literal retrieval, where attenuation is close to a local
write-direction regime, the predictor is quantitatively strong: $R^2{=}0.888$ (Pythia),
$0.601$ (Qwen2.5), and $0.867$ (Qwen3.5). Semantic retrieval is more revealing when raw
$R^2$ is read together with directional diagnostics (Table~\ref{tab:mechanistic-diagnostics}):
Pythia and Qwen2.5 have low raw $R^2$ but preserve sign and rank signal, while Qwen3.5
already has higher raw $R^2$. The LongBench retrieval-QA probe (\S\ref{sec:longbench})
recovers raw $R^2{=}0.796$ (Qwen2.5) and $0.791$ (Qwen3.5) under natural QA, showing that
the controlled-semantic amplitude gap is protocol-specific rather than a failure of the
directional mechanism.

\begin{table}[h]\centering\small
\caption{\textbf{Directional-derivative diagnostics separate direction from amplitude.}
The local directional predictor uses $D_l^{\mathrm{pred}}(\alpha)=(1-\alpha)s_l$ on attenuation gains. Literal
retrieval is close to the zero-parameter amplitude model. Semantic retrieval can lower raw
$R^2$ while preserving directional signal; majority, balanced accuracy, and MCC show that
semantic sign agreement is not a class-imbalance artifact.}
\label{tab:mechanistic-diagnostics}
\begin{tabular}{llrrrrrr}
\toprule
Model & Suite & raw $R^2$ & Pearson & Sign & Majority & Bal. & MCC \\\midrule
Pythia-1B & literal & 0.888 & -- & 0.917 & -- & -- & -- \\
Qwen2.5-7B & literal & 0.601 & -- & 0.845 & -- & -- & -- \\
Qwen3.5-9B & literal & 0.867 & -- & 0.906 & -- & -- & -- \\
Pythia-1B & semantic & 0.139 & 0.582 & 0.792 & 0.542 & 0.773 & 0.628 \\
Qwen2.5-7B & semantic & 0.074 & 0.770 & 0.821 & 0.643 & 0.861 & 0.694 \\
Qwen3.5-9B & semantic & 0.639 & 0.820 & 0.792 & 0.521 & 0.787 & 0.592 \\\bottomrule
\end{tabular}
\end{table}

\paragraph{Semantic retrieval separates direction from propagation.} The zero-parameter
predictor tests two things at once: whether $s_l$ gets the \emph{direction/order} of
damage right, and whether its uncalibrated scale matches the observed damage amplitude.
Semantic retrieval often preserves the first while weakening the second. In Qwen2.5
semantic retrieval, raw $R^2$ is $0.074$, but Pearson correlation is $0.770$, sign accuracy
is $0.821$, balanced sign accuracy is $0.861$, and MCC is $0.694$. This turns semantic
retrieval into a mechanism probe: it changes the task gradient and often flips the same
layer's direction, while the resulting damage magnitude depends more on how that write
propagates through later layers and depth structure. The signal-graph diagnostic in
Appendix~\ref{app:signalgraph} supports this interpretation: it is used as a propagation
check, while $s_l$ remains the sign anchor for the monotone roles.

\paragraph{Magnitude vs direction.} Separately, a 33-dim residual-magnitude feature model
is useful for predicting damage magnitude, but it does not explain the taxonomy. In
three-class leave-one-model-out role prediction (calibrated/suppressor/amplifier), relative norm
alone gives balanced accuracy $0.20$ against $0.33$ balanced chance; adding depth improves
to $0.32$, and residual norm/cosine/depth features reach $0.33$. Magnitude carries partial
signal, but it is unsigned and cannot tell suppressor from amplifier. The role is set by
the \emph{direction} ($s_l$), which magnitude alone cannot recover.

\paragraph{Propagation and depth diagnostics.} Block-level attenuation and the
forward-only signal graph support the same qualitative interpretation: attenuation damage
also depends on downstream propagation and depth structure. These diagnostics are not used
to define endpoint roles or the $s_l$ mechanism, so we keep the details in
Appendices~\ref{app:block} and~\ref{app:signalgraph}.

\subsection{Transfer: LongBench retrieval-QA preserves the signed surface}
\label{sec:longbench}

The controlled literal/semantic suites isolate retrieval from reasoning and domain
knowledge. We next ask how the same signed-surface analysis behaves under a more natural
QA protocol, and repeat native FFN scaling on the safety-filtered LongBench retrieval-QA
probe for the two Qwen
models. The grids are complete and directly auditable from predictions: Qwen2.5-7B has
253 conditions and 50{,}600 prediction rows; Qwen3.5-9B has 289 conditions and 57{,}800
prediction rows. Rebuilding the summaries from predictions recovers $D_l(\alpha)$ to
floating-point precision (maximum absolute margin-damage differences $2.2{\times}10^{-16}$
and $9.7{\times}10^{-17}$).

\begin{figure}[h]
\centering
\includegraphics[width=\linewidth]{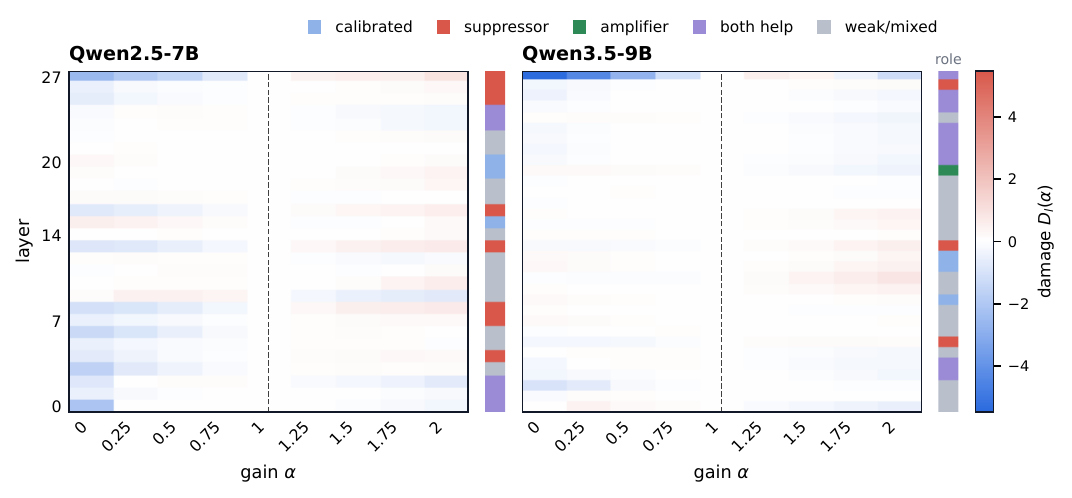}
\caption{\textbf{The safety-filtered LongBench retrieval-QA probe also has signed FFN response surfaces.}
Heatmaps show $D_l(\alpha)$ for Qwen2.5-7B and Qwen3.5-9B across every layer and gain on
candidate-scored Qasper and MultiFieldQA-en examples. Signed structure persists under the
natural QA probe, while the final-layer role can shift across protocols.}
\label{fig:longbench}
\end{figure}

Figure~\ref{fig:longbench} shows that the signed surface persists: gain changes do not
produce uniform degradation, and both Qwen models contain multiple endpoint types. Under
the same $|D_l(\alpha)|{>}0.1$ endpoint threshold, Qwen2.5 has 3 calibrated layers,
8 suppressors, 0 amplifiers, 5 both-ends-help layers, and 12 weak/mixed layers; Qwen3.5
has 3 calibrated layers, 3 suppressors, 1 amplifier, 9 both-ends-help layers, and
16 weak/mixed layers (Table~\ref{tab:longbench}). Thus the naturalistic probe does
not merely reproduce the controlled taxonomy counts; it shows the same measurement
recovering a signed layer-role surface under a natural QA protocol.

The directional-derivative mechanism also transfers. Using LongBench backward diagnostics and
attenuation gains $\alpha\in\{0.25,0.5,0.75\}$, the zero-parameter predictor
$D_l^{\mathrm{pred}}(\alpha)=(1-\alpha)s_l$ explains damage with raw $R^2{=}0.796$ for Qwen2.5 and
$0.791$ for Qwen3.5; Pearson correlations are $0.903$ and $0.949$, and balanced sign
accuracies are $0.857$ and $0.802$. The per-subtask breakdown is similarly strong
(Appendix Table~\ref{tab:longbench-subtask}): Qasper and MultiFieldQA-en each have raw
$R^2{>}0.75$ and Pearson correlation above $0.89$. The weakest sign row is Qwen3.5/Qasper,
where sign accuracy is $0.708$ against a $0.594$ majority baseline (balanced accuracy
$0.710$). Unlike the lower-$R^2$ controlled
semantic cases, this probe does not show the same direction-versus-amplitude split for the
Qwen models: the local directional derivative is quantitatively predictive of attenuation damage.

\begin{table}[h]\centering\small
\caption{\textbf{LongBench retrieval-QA transfer summary.}
Pooled Qasper and MultiFieldQA-en results for the two Qwen models.}
\label{tab:longbench}
\begin{tabular}{lrrrrrlrrr}
\toprule
Model & Calib. & Supp. & Amp. & Both & Weak & Final & $R^2$ & Pearson & Sign/Bal. \\\midrule
Qwen2.5-7B & 3 & 8 & 0 & 5 & 12 & supp. & 0.796 & 0.903 & 0.845/0.857 \\
Qwen3.5-9B & 3 & 3 & 1 & 9 & 16 & both & 0.791 & 0.949 & 0.802/0.802 \\
\bottomrule
\end{tabular}
\end{table}

Both-help is reported explicitly here because it is frequent under the natural QA protocol:
5/28 Qwen2.5 layers and 9/32 Qwen3.5 layers in the pooled rows. This is a protocol-shift
role, not a replacement for the monotone mechanism claim; the suppressor/amplifier analysis
continues to center on the two single-direction roles and their $s_l$ signs.

The final-layer shift is informative. In the controlled retrieval protocol, the final FFN is
usually a readout-adjacent suppressor. In the LongBench retrieval-QA probe, Qwen2.5 retains
that aggregate pattern ($D_l(0){=}{-}2.60$, $D_l(2){=}+0.90$), while Qwen3.5 moves into a
both-ends-help role ($D_l(0){=}{-}5.47$, $D_l(2){=}{-}1.25$): both skipping and doubling the final
FFN improve the margin. This supports the paper's main framing. FFN roles are
task-conditioned signed-write roles; final-layer suppression is a strong controlled
retrieval phenomenon, and natural QA can move the same architectural position into a
different signed role.

Finally, the signed surface supports a held-out policy sanity check, reported in full in
Appendix~\ref{app:longbenchpolicy}. We select top suppressor layers on a calibration split
and evaluate only on a disjoint LongBench split. With $k{=}3$ layers and
$\alpha{=}0.5$, attenuating the selected suppressors improves held-out margin in both Qwen
models (mean $+3.01$), exceeding random layer sets ($+0.13$) and relative-norm-matched
controls ($-0.14$). This is not proposed as an optimizer; it is a compact check that the
taxonomy identifies causally useful layer directions.

\clearpage

\subsection{Minimal robustness and confound controls}
\label{sec:robust}

The role taxonomy is defined by signs of $D_l(0)$ and $D_l(2)$ with a
$|D_l(\alpha)|{<}0.1$ weak band. We therefore audit its three main degrees of freedom:
threshold choice, seed variation, and FFN update size
(\texttt{scripts/robustness\_checks\_20260621.py}).

\paragraph{Threshold sensitivity.} The exact counts move when the weak band changes, as
they should, but the taxonomy does not collapse into one response type. For Qwen2.5-7B
literal, the number of calibrated layers is 18 at thresholds 0.05 and 0.1, and remains 13 at
0.25. For Qwen3.5-9B literal, the exact count shifts more (calibrated 8/6/3 at
0.05/0.1/0.25), but suppressor and amplifier classes persist. Thus the threshold affects
borderline labels, not the qualitative finding that calibrated, suppressive, amplifying,
and weak/mixed roles coexist.

\paragraph{Seed robustness (leave-one-seed-out).} For the three multi-seed models we
recompute every layer's role under each of three leave-one-seed-out folds. $136/152$ layers
($89\%$) retain the same role label across all three folds, and all 6 end layers retain the
suppressor label across folds. The taxonomy is therefore not a single-seed artifact.

\paragraph{Bootstrap confidence intervals.} Seed-bootstrap CIs ($B{=}2000$) on end-layer
damage values are narrow (Qwen3.5-9B literal: $D_l(0)=-3.49\,[-3.70,-3.31]$,
$D_l(2)=+3.10\,[+3.02,+3.15]$); the weakest endpoint is Qwen2.5-7B semantic
$\alpha{=}0$ ($-0.17\,[-0.30,-0.08]$, close to the weak band). Accordingly, the end-layer
claim is a qualitative monotone-suppressor pattern, not equal endpoint strength in every
suite. The role-switch rate is $60\%$ with layer-bootstrap 95\% CI
$[50\%,69\%]$.

\begin{figure}[h]
\centering
\includegraphics[width=0.90\linewidth]{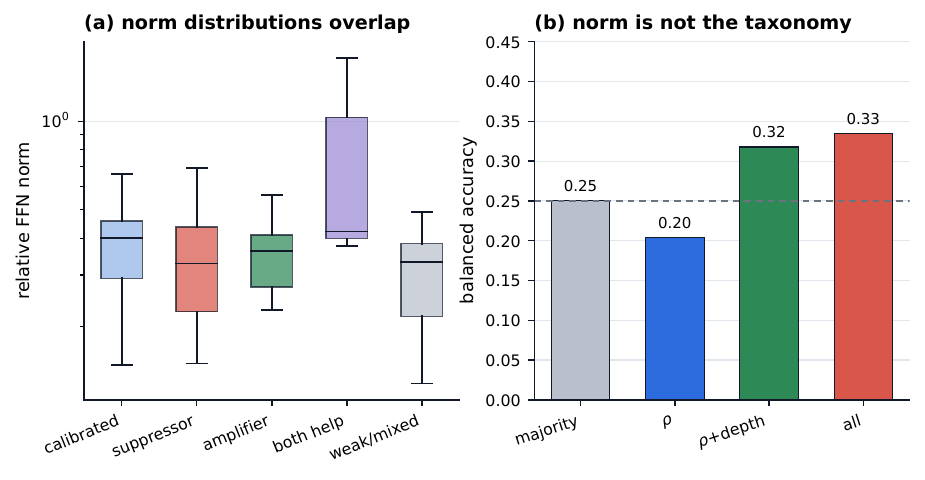}
\caption{\textbf{Residual size is a partial signal, not a taxonomy explanation.}
\textbf{(a)} FFN relative norm overlaps across endpoint taxonomy types.
\textbf{(b)} Leave-one-model-out taxonomy prediction from residual size/depth features.
The classifier uses the three signed roles and omits weak/mixed layers; relative norm alone
is below balanced chance, while adding depth and residual/cosine features remains weak.}
\label{fig:confound}
\end{figure}

\paragraph{Residual-size controls.} Relative FFN norm, depth, and residual geometry do not
explain the taxonomy by themselves. Figure~\ref{fig:confound} gives the minimal control.
Relative FFN norm overlaps across role classes (median calibrated $0.40$,
suppressor $0.33$, amplifier $0.36$; IQRs overlap substantially). In three-class
leave-one-model-out role prediction, relative norm alone gives balanced accuracy $0.20$
against $0.33$ balanced chance; adding depth improves to $0.32$, and residual
norm/cosine/depth features reach $0.33$. Residual size and depth carry partial signal, but
they do not determine the signed role. The direction ($s_l$), which magnitude cannot
recover, is what predicts the monotone roles. Appendix~\ref{app:fixednorm} adds a
fixed-norm rotation sanity check on Qwen2.5: rotating writes with or against the local
retrieval-gradient direction while preserving write and residual norms gives true-gradient
win rates of $0.997$ (literal/main) and $1.000$ (semantic/supplementary), while random
tangent rotations stay near chance ($0.513$ and $0.486$).

% =============================================================================
\section{Discussion}
% =============================================================================
\label{sec:discussion}

\paragraph{Design consequence.} Memory-decoupling architectures isolate storage from dense
computation; Memory Layers, DeepSeek Engram, MemoryLLM, and STEM are examples. Our results
do not show that all such designs are at risk. They identify a diagnostic axis for
retrieval settings: when an architecture replaces or factors out FFN memory-like
computation, lookup accuracy alone may miss whether the replacement write pushes the
residual state in the task-relevant direction. The directional-derivative score $s_l$
provides a zero-parameter post-hoc check: after replacement or decoupling, measure whether
the module's write still aligns with the retrieval gradient.
The LongBench policy sanity check makes the diagnostic actionable within our setting:
attenuating top
suppressor layers selected on a calibration split improves held-out margin by $+3.01$ on
average at $k{=}3,\alpha{=}0.5$, compared with $+0.13$ for random layer sets and $-0.14$
for relative-norm-matched controls.

\paragraph{What the end-layer suppressor means.} In the controlled retrieval protocol, the
end layer sits immediately before the output head; there is no later attention for it to
shape. Its suppressor signature is
therefore best read as a \emph{readout-adjacent} effect: the end-layer FFN write injects or
amplifies information that competes with the retrieval decision at output scoring --- e.g.\
prior-token continuations or majority-class priors --- so removing it clarifies the
retrieved signal and amplifying it drowns it. This is the same signed-write question at the
output boundary: non-final layers can change what later attention selects, while the final
layer changes what the output head scores.

\paragraph{What LongBench changes.} The safety-filtered LongBench retrieval-QA probe makes this separation
concrete. It preserves signed FFN response surfaces and strong $s_l$ predictivity in a
more natural QA setting, while changing the readout-adjacent role of Qwen3.5's final layer
to both-ends-help. This separation is useful: the central claim is a signed,
task-conditioned native-write taxonomy, and final-layer suppression is one prominent
controlled-retrieval role within that taxonomy. Naturalistic QA can change which direction a
readout-adjacent FFN write pushes the retrieval margin, while the signed-surface and
directional-derivative structure remains visible.

\paragraph{Why the role switches between tasks (Q2).} Literal retrieval needs attention to
surface a verbatim needle; semantic retrieval needs it to surface an associated concept.
The same FFN write can rotate from aiding one to suppressing the other: in
Figure~\ref{fig:sl}, several layers have $s_l$ of opposite sign across suites. This is the
mechanism behind the 60\% role switch and the core evidence that FFN's role is
task-conditioned rather than a static property of stored knowledge.

\paragraph{Claim calibration.} The evidence supports a strong but specific claim: in
long-context retrieval, native FFN writes have signed, task-conditioned effects on retrieval margin.
Direction is the decisive signal for the two monotone roles; residual size, depth, and
propagation features shape attenuation magnitude and explain why the full response surface
is richer than a single sign. The mechanism anchor (Q3) is verified across both
residual-composition families, with SmolLM2 providing cross-protocol validation of the
Pythia/Qwen protocol. The role landscape is depth-structured and mixed in the middle
layers, so the paper reports a signed layer-role distribution rather than a depth-only rule.

% =============================================================================
\section{Limitations}
% =============================================================================
\label{sec:limitations}
\paragraph{Pythia's parallel residual.} Pythia computes
$h_{l+1}{=}h_l{+}\mathrm{Attn}{+}\mathrm{FFN}$ with $\mathrm{FFN}$ reading $h_l$ (not $u_l$).
The native-write intervention is valid in both residual-composition families, but the
``reshapes the next attention's input'' framing is cleanest on the three sequential models.
Pythia is included as a cross-family check.

\paragraph{Backward scope.} The mechanism anchor (Q3) uses per-layer backward data (bf16
full-alpha) on the three main-protocol models and sampled backward anchor layers on
SmolLM2. Qwen semantic backward diagnostics exclude the short-copy control and average only
the semantic tasks. The $s_l\leftrightarrow$role mapping is therefore checked across both
residual-composition families and a range of model scales (1--9B), but the local directional
amplitude model is used as a task-local diagnostic rather than a universal cross-model
calibration rule.

\paragraph{Training dynamics.} We show that trained FFN writes have task-conditioned signed effects on retrieval. The current experiments characterize the trained mechanism; a
checkpoint study would test how these directions form during training.

\paragraph{Vector-subspace decomposition.} This paper establishes a signed causal surface
with native scaling interventions and robustness controls. Decomposing individual FFN
writes into finer vector subspaces is a complementary analysis, not a prerequisite for the
claims here.

% =============================================================================
\section{Conclusion}
% =============================================================================
\label{sec:conclusion}
Taken together, the results show that the role of FFNs in LLMs is more than memory. FFNs do
not only store or retrieve associations; in a native Transformer, their outputs are written
inline into the residual stream that later attention layers and the output head consume. In
long-context retrieval, this gives the native FFN write a signed role: it can be calibrated,
amplifying, or suppressive depending on layer, task, and direction.

Our theoretical and empirical analyses support this view. Native FFN scaling isolates the
model's own residual write and reveals signed, layer-specific response surfaces. Across the
controlled main protocol and SmolLM2 cross-protocol validation, final-layer writes are
suppressors in 7 of 8 model--suite pairs, 60\% of layers switch role between literal and
semantic retrieval (95\% CI [50\%, 69\%]), and the monotone roles are anchored by the local
directional derivative: suppressor $s_l{<}0$ in 34/35 cases and amplifier $s_l{>}0$ in 18/18 cases. A
safety-filtered LongBench retrieval-QA probe preserves the signed surface on two Qwen
models and gives strong first-order prediction ($R^2{=}0.796$ / $0.791$), while also
showing that final-layer roles can change under naturalistic protocol shift.

For architectures that decouple, sparsify, or replace FFN parameters, the implication is a
diagnostic one rather than a universal design rule. In retrieval settings, lookup accuracy
should be complemented by a check on the contextual role of the replacement write: where it
enters the residual stream, how later attention and the output head consume it, and whether
its direction pushes the retrieval state toward or away from the current task objective.
The storage view explains localization and editing of facts; the signed-write view adds a
directional question to ask when FFN memory-like computation is factored out of the native
Transformer.

\appendix

\section{Taxonomy Threshold Sweep}
\label{app:thresholds}

The main taxonomy uses the pre-specified weak band $|D_l(\alpha)|<0.1$. Table~\ref{tab:thresholds}
reports the complete threshold sweep used in the robustness audit. The threshold changes
borderline labels, especially for weak/mixed layers, but does not collapse the signed
taxonomy into a single response type.

\begin{table}[h]\centering\scriptsize
\caption{\textbf{Endpoint-taxonomy threshold sweep.} Counts are recomputed from the same
seed-averaged endpoints using weak bands 0.05, 0.10, and 0.25.}
\label{tab:thresholds}
\begin{tabular}{llrrrrrr}
\toprule
Thr. & Model/Suite & $n$ & Calib. & Supp. & Amp. & Both & Weak \\\midrule
0.05 & Pythia literal & 16 & 3 & 6 & 3 & 2 & 2 \\
0.05 & Pythia semantic & 16 & 10 & 3 & 1 & 0 & 2 \\
0.05 & Qwen2.5 literal & 28 & 18 & 4 & 4 & 1 & 1 \\
0.05 & Qwen2.5 semantic & 28 & 16 & 5 & 4 & 0 & 3 \\
0.05 & Qwen3.5 literal & 32 & 8 & 10 & 8 & 0 & 6 \\
0.05 & Qwen3.5 semantic & 32 & 9 & 10 & 2 & 1 & 10 \\
0.10 & Pythia literal & 16 & 2 & 5 & 2 & 2 & 5 \\
0.10 & Pythia semantic & 16 & 8 & 3 & 0 & 0 & 5 \\
0.10 & Qwen2.5 literal & 28 & 18 & 4 & 4 & 0 & 2 \\
0.10 & Qwen2.5 semantic & 28 & 14 & 4 & 3 & 0 & 7 \\
0.10 & Qwen3.5 literal & 32 & 6 & 8 & 6 & 0 & 12 \\
0.10 & Qwen3.5 semantic & 32 & 9 & 9 & 2 & 1 & 11 \\
0.25 & Pythia literal & 16 & 1 & 4 & 0 & 1 & 10 \\
0.25 & Pythia semantic & 16 & 7 & 3 & 0 & 0 & 6 \\
0.25 & Qwen2.5 literal & 28 & 13 & 4 & 3 & 0 & 8 \\
0.25 & Qwen2.5 semantic & 28 & 11 & 2 & 2 & 0 & 13 \\
0.25 & Qwen3.5 literal & 32 & 3 & 4 & 4 & 0 & 21 \\
0.25 & Qwen3.5 semantic & 32 & 4 & 4 & 0 & 0 & 24 \\\bottomrule
\end{tabular}
\end{table}

\section{Four-Block Attenuation Diagnostic}
\label{app:block}

\begin{figure}[h]
\centering
\includegraphics[width=\linewidth]{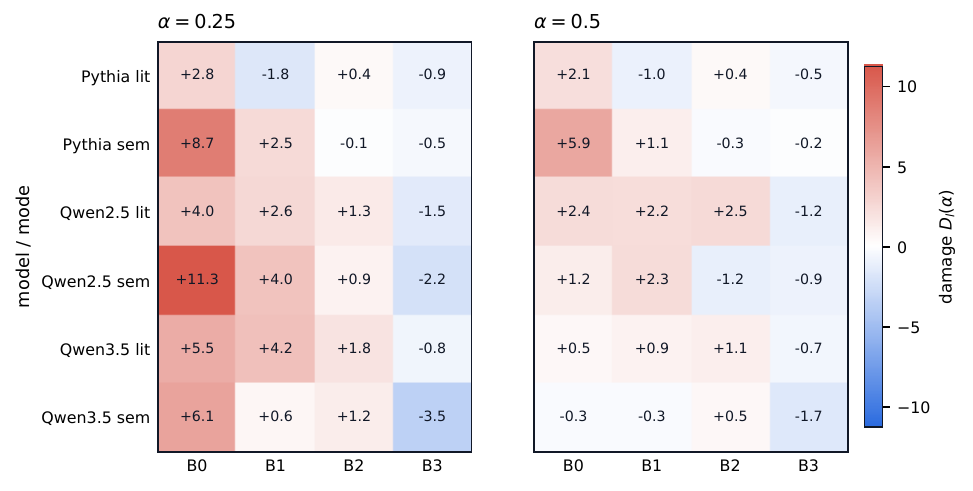}
\caption{\textbf{Four-block attenuation exposes coarse depth roles.}
Each network is split into four contiguous FFN blocks. Cells show damage from scaling all
FFN writes in one block to $\alpha{=}0.25$ or $\alpha{=}0.5$. Positive values mean block
attenuation hurts retrieval; negative values mean it helps.}
\label{fig:block}
\end{figure}

The single-layer taxonomy could be read as isolated per-layer effects. As a coarse
propagation check, we split each network into four contiguous FFN blocks by depth and
scale all FFN writes in one block at a time. Figure~\ref{fig:block} shows that the signed
depth structure persists at block scale. This diagnostic is not used to define endpoint
roles or to replace the local $s_l$ sign test.

Weakening the \emph{earliest} block at $\alpha{=}0.25$
is the most damaging block intervention in all six main (model,suite) pairs:
damage $+2.80$ (Pythia literal), $+8.68$ (Pythia semantic), $+4.02$ (Qwen2.5
literal), $+11.26$ (Qwen2.5 semantic), $+5.51$ (Qwen3.5 literal), and $+6.09$ (Qwen3.5
semantic). Weakening the \emph{last} block has the opposite sign in the same six pairs:
$-0.92$, $-0.46$, $-1.54$, $-2.22$, $-0.79$, and $-3.47$. Because a block contains several
FFNs (including the final suppressor in the last quarter), this is not a per-layer
attribution; it is a coarse confirmation that signed depth structure survives when
contiguous depth segments are perturbed together. SmolLM2 supplies the cross-protocol
validation: weakening its earliest block at $\alpha{=}0.25$ hurts literal and semantic
retrieval ($+3.83$ and $+7.57$), while weakening its last block helps literal retrieval
($-1.46$) and hurts semantic retrieval ($+1.26$), again showing signed depth structure
rather than one uniform FFN effect.

\section{Signal-Graph Propagation Diagnostic}
\label{app:signalgraph}

The signal-graph diagnostic is a forward-only propagation check. It predicts attenuation
responses from FFN update magnitude and residual-stream propagation features (norms,
ratios, cosines crossed with dose), fit with seed leave-one-out over the three
main-protocol models. The source predictions use the delta-margin sign convention, but
$R^2$ is invariant under the global sign convention used for $D_l(\alpha)$ in the paper.

\begin{table}[h]\centering\small
\caption{\textbf{Signal-graph propagation diagnostic.} These rows support the claim that
attenuation effects follow residual-stream propagation structure. They are not used to
define endpoint roles and are not a replacement for the local $s_l$ sign test.}
\label{tab:signalgraph}
\begin{tabular}{llrr}
\toprule
Diagnostic & Scope & $R^2$ & Sign acc. \\\midrule
Signal graph & three main-protocol models, $\alpha<1$ & 0.948 & 0.853 \\
Signal graph & Qwen2.5 only, $\alpha<1$ & 0.904 & 0.883 \\
Signal graph & Qwen3.5 only, $\alpha<1$ & 0.798 & 0.780 \\
Signal graph & literal responses, $\alpha<1$ & 0.902 & 0.865 \\
Signal graph & semantic responses, $\alpha<1$ & 0.972 & 0.841 \\\bottomrule
\end{tabular}
\end{table}

\section{LongBench Policy and Gradient-Path Sanity}
\label{app:longbenchpolicy}

\begin{table}[h]\centering\small
\caption{\textbf{Per-subtask LongBench retrieval-QA breakdown.} These rows support the
pooled LongBench table in the main text and show that the transfer result is not driven by
only one subtask.}
\label{tab:longbench-subtask}
\begin{tabular}{llrrrr}
\toprule
Model & Subtask & $R^2$ & Pearson & Sign & Bal. \\\midrule
Qwen2.5-7B & MultiFieldQA-en & 0.805 & 0.940 & 0.869 & 0.896 \\
Qwen2.5-7B & Qasper & 0.752 & 0.892 & 0.821 & 0.824 \\
Qwen3.5-9B & MultiFieldQA-en & 0.781 & 0.954 & 0.896 & 0.904 \\
Qwen3.5-9B & Qasper & 0.808 & 0.945 & 0.708 & 0.710 \\\bottomrule
\end{tabular}
\end{table}

\section{Fixed-Norm Directional Rotation Sanity}
\label{app:fixednorm}

The residual-size controls in the main text use scalar diagnostics. As a more direct
sanity check, we also run a Qwen2.5 fixed-norm rotation pilot: for selected layers, we
rotate the native write in the positive or negative local gradient direction while
renormalizing the write and resulting residual state to preserve their original norms. This
is not a replacement for native scaling, because it constructs a tangent perturbation
rather than varying the model's own gain. It is a targeted test of whether direction alone
can move retrieval margin when norm is held fixed.

\begin{table}[h]\centering\small
\caption{\textbf{Fixed-norm directional rotation sanity check.} Positive gap means the
positive-gradient rotation improves margin relative to the negative-gradient rotation.
The maximum write/residual norm-ratio deviation from 1 is below $2{\times}10^{-6}$ in all
rows.}
\label{tab:fixednorm}
\begin{tabular}{llrrrr}
\toprule
Suite & Rotation & Rows & Mean gap & Win rate & Pred./obs. corr. \\\midrule
Main literal & true gradient & 720 & 41.33 & 0.997 & 0.990 \\
Main literal & random tangent & 720 & 0.01 & 0.513 & -- \\
Semantic supp. & true gradient & 720 & 82.51 & 1.000 & 0.962 \\
Semantic supp. & random tangent & 720 & -0.08 & 0.486 & -- \\\bottomrule
\end{tabular}
\end{table}

The true-gradient direction wins almost every paired comparison, while random tangent
directions stay near chance. Because the write and residual norms are effectively fixed,
this supports the main claim that the sign of the FFN effect is directional rather than a
simple consequence of update size.

\paragraph{Directional-derivative path sanity.} We validate the LongBench directional-derivative path with
finite differences on five anchor layers per Qwen model. The finite-difference direction
and the autograd alpha-gradient direction agree in sign on all sampled anchors
(Table~\ref{tab:fd_sanity}). The direct dot product and the recorded alpha-gradient are
also close, so the sanity check supports the implementation of $s_l$ rather than adding a
separate mechanistic claim.

\begin{table}[h]\centering\small
\caption{\textbf{LongBench finite-difference sanity check.} Each row uses one Qasper sample
and five anchor layers. FD sign compares finite differences with the recorded
alpha-gradient direction; dot-alpha sign compares the direct $\Delta F_l^\top\nabla M$ dot
product with the alpha-gradient direction.}
\label{tab:fd_sanity}
\begin{tabular}{llrrr}
\toprule
Model & Anchor layers & FD sign & Dot-alpha sign & Mean $|\epsilon_{\mathrm{dot},\alpha}|$ \\\midrule
Qwen2.5-7B & 0, 7, 14, 21, 27 & 5/5 & 5/5 & 0.024 \\
Qwen3.5-9B & 0, 8, 16, 24, 31 & 5/5 & 5/5 & 0.052 \\\bottomrule
\end{tabular}
\end{table}

\paragraph{Held-out policy controls.} Table~\ref{tab:policy_full} gives the full LongBench
held-out policy sanity table. Top suppressor layers are selected from the single-layer
margin-damage screen on a calibration split; random and relative-norm-matched controls use
the same $k$ and policy alpha. All values are evaluated on a disjoint 100-example
evaluation split. The main text reports the conservative attenuation rows
($k{=}3,\alpha{=}0.5$). The ablation rows ($\alpha{=}0$) preserve the same ordering but
are not used as the main policy claim because they are a stronger intervention.

\begin{table}[h]\centering\small
\caption{\textbf{Held-out LongBench FFN-policy comparison.} Positive values are evaluation-split
margin gains over baseline.}
\label{tab:policy_full}
\begin{tabular}{llrrrr}
\toprule
Model & $k$ & $\alpha$ & Top suppressors & Random & Norm matched \\\midrule
Qwen2.5-7B & 3 & 0.5 & 2.72 & 0.30 & -0.33 \\
Qwen2.5-7B & 3 & 0.0 & 4.48 & 0.38 & -8.75 \\
Qwen2.5-7B & 5 & 0.5 & 3.75 & 1.90 & -0.24 \\
Qwen2.5-7B & 5 & 0.0 & 4.77 & 3.21 & -14.13 \\
Qwen3.5-9B & 3 & 0.5 & 3.30 & -0.04 & 0.05 \\
Qwen3.5-9B & 3 & 0.0 & 7.66 & 0.09 & 0.54 \\
Qwen3.5-9B & 5 & 0.5 & 3.38 & 0.12 & 0.09 \\
Qwen3.5-9B & 5 & 0.0 & 7.96 & 0.56 & 1.25 \\\bottomrule
\end{tabular}
\end{table}

\bibliographystyle{plainnat}
\bibliography{refs}

\end{document}